\pdfoutput=1

\documentclass[11pt]{article}
\usepackage{graphicx}
\usepackage[table,xcdraw]{xcolor}
\usepackage{multirow}
\usepackage{enumitem}
\usepackage{pdfpages}
\usepackage{amsmath}

\usepackage{acl}

\usepackage{times}
\usepackage{latexsym}

\usepackage[T1]{fontenc}

\usepackage[utf8]{inputenc}

\usepackage{microtype}

\title{LMs stand their Ground: \\ Investigating the Effect of Embodiment in Figurative Language Interpretation by Language Models}

\author{Philipp Wicke \\
  Ludwig-Maximilians-University (LMU) \\
  Institute for Information and Language Processing (CIS) \\
   Munich Center for Machine Learning (MCML)  \\
  \texttt{pwicke@cis.uni-muenchen.de} }

\begin{document}
\maketitle
\begin{abstract}
Figurative language is a challenge for language models since its interpretation is based on the use of words in a way that deviates from their conventional order and meaning. Yet, humans can easily understand and interpret metaphors, similes or idioms as they can be derived from embodied metaphors. Language is a proxy for embodiment and if a metaphor is conventional and lexicalised, it becomes easier for a system without a body to make sense of embodied concepts. Yet, the intricate relation between embodiment and features such as concreteness or age of acquisition has not been studied in the context of figurative language interpretation concerning language models. Hence, the presented study shows how larger language models perform better at interpreting metaphoric sentences when the action of the metaphorical sentence is more embodied. The analysis rules out multicollinearity with other features (e.g. word length or concreteness) and provides initial evidence that larger language models conceptualise embodied concepts to a degree that facilitates figurative language understanding.
\end{abstract}

\section{Introduction}

Infants acquire their first conceptual building blocks by observation and manipulation in the physical world. These primary building blocks enable them to make sense of their perceptions \cite{mandler2014defining}. In return, their embodiment defines the capabilities with which they can explore and understand the world. The early conceptual system is built from spatial schemas, which enables early word understanding \cite{mandler1992build}. These so-called \textit{Image Schemas} are recurring cognitive structures shaped by physical interaction with the environment. They emerge from bodily experience and motivate subsequent conceptual metaphor mappings \cite{johnson2013body}. The metaphorical mapping is visible in our everyday language whenever we use figurative language. For example, if we say that \textit{she \textcolor{orange}{dances} like a turtle}, that is to say, that \textit{she dances poorly}. The metaphor in this phrase is readily interpreted by humans, who would favour the interpretation of \textit{dances poorly} over \textit{dances well}. The \textit{turtle dance} example employs a conceptual mapping in which the \textit{turtle} provides the source domain for the attributes \textit{slow} and \textit{rigid}, which turn the \textit{dance} target domain into \textit{poorly dancing}. This mapping draws from the human, bodily experience of dancing and therefore enables interpretation.

For a language model (LM), the understanding of figurative language is a great challenge \cite{liu2022testing}. By nature of their digital implementation as computer algorithms, LMs are non-embodied and do not ground their conceptualisation by physical interaction with the environment. Instead, LMs learn statistical features of language by deep learning vast amounts of data \cite{vaswani2017attention}. Whether these learned statistical features allow LMs to mirror or copy natural language understanding (NLU) is subject to discussion \cite{zhang2022paradox}. Moreover, \citet{tamari2020language} suggest an \textit{embodied} language understanding paradigm for LMs can benefit NLU systems through grounding by metaphoric inference.

One can argue that most embodied metaphors are heavily conventional (e.g. UP IS GOOD, DOWN IS BAD, KNOWLEDGE IS LIGHT, IGNORANCE IS DARKNESS) and as such, they are lexicalised in a language without an inherent need to understand their bodily basis. This lexicalisation should allow LMs to conceptualise and interpret them correctly and more robustly than less conventional metaphors. Conventionality relates to word frequency and age of acquisition (AoA), i.e. more frequent words and words that are acquired early in life are more conventional. We argue that embodiment has a measurable effect on the interpretation of metaphors that differs from the effect of other linguistic features. Moreover, we investigate whether an interpretation of figurative language with more embodied concepts is easier for LMs. Analogously, we investigate conflating factors such as the AoA, word frequency, concreteness and word length. The relation between embodiment and LMs' ability to interpret figurative language has not yet been investigated and is the key contribution of this research. 

The following Section (\ref{sec:lit}) starts with a review of language model abilities, more specifically figurative language interpretation abilities. The review identifies a suitable data set for our experiment and describes its formation in Section \ref{sec:stats}. We use a subset of the \textit{Fig-QA} data set \cite{liu2022testing}, a Winograd-style figurative language understanding task, and correlate the performance of various LMs concerning the degree of embodiment of the metaphorical actions that the LMs are tasked to interpret. In Section \ref{sec:results}, we identify that models, that can reach a certain performance on our \textit{Fig-QA} subset, shows a significant and positive correlation between the rating of the embodiment of the action involved in the metaphorical phrase and the model's ability to interpret the metaphor correctly. An in-depth analysis of additional features, such as the AoA, word length or frequency, does not indicate multicollinearity among those features. In Section \ref{sec:concl} we conclude that the degree of embodiment of the action within the metaphoric phrase is a predictor of the LMs' ability to correctly interpret the figurative language. Lastly, we discuss the limitations and broader implications of the work.

\section{Related Works}
\label{sec:lit}

\subsection{Language Model Abilities}

The presented work investigates the zero-shot capabilities of LMs of different types and sizes. Arguably, LMs' capabilities to solve language-based tasks, which they have not been trained on, are an emerging property of their complexity and large-scale statistical representation of language. It is a property that makes them unsupervised multitask learners \cite{radford2019language, brown2020language}. Despite task-agnostic pre-training and a task-agnostic architecture, LMs can perform various NLP tasks without seeing a single example of the task, albeit with mixed results \cite{srivastava2022beyond}. This raises the question of whether language models mirror the human conceptual understanding encoded in language or whether they ``only'' learn statistical features from the underlying training distribution, allowing them to generalise and convincingly solve previously unseen tasks.

Several works have tried to assess to what extent LMs are capable to perform more complex NLP tasks (e.g. logical reasoning or metaphoric inference). For example, \citet{zhang2022paradox} investigate the logical reasoning capabilities of BERT \cite{devlin2018bert}. For this, the authors define a simplistic problem space for logical reasoning and show that BERT learns statistical features from its training distribution, but fails to generalise when presented with other distributions and drops in performance. According to the authors, this implies that BERT does not emulate a correct reasoning function in the same way that humans would conceptualise the problem. Similarly, \citet{sanyal2022robustlr} evaluate whether the RoBERTa model \cite{liu2019roberta} or the T5 model \cite{raffel2020exploring} can perform logical reasoning by understanding implicit logical semantics. The authors test the models on various logical reasoning data sets whilst introducing minimal logical edits to their rule base. Consequently, \citet{sanyal2022robustlr} show that LMs, even when fine-tuned on logical reasoning, do not sufficiently learn the semantics of some logical operators. \citet{han2022folio} present a diverse data set for reasoning in natural language. An evaluation of the GPT-3 model \cite{brown2020language} on their data set shows a performance that is only slightly better than random. This indicates that there is a fundamental gap between human reasoning and LM reasoning and their conceptualisation capabilities. Yet, language models have demonstrated emergent abilities \cite{wei2022emergent}, encompassing enhanced skills and capabilities that are absent in smaller language models. Such abilities cannot be accurately predicted by extrapolating the performance of smaller models. Consequently, investigating the influence of model size on different tasks becomes imperative in comprehending the potentials and constraints of smaller and large language models.

The related works show that, although LMs seem to mirror an aspect of reasoning, e.g. logical reasoning, a closer look at the underlying conceptualisation of these abilities can reveal they are not robust and fail to mirror deeper semantics. Both logical reasoning and figurative language interpretation require an understanding of relationships between words and concepts and the ability to make inferences based on that understanding. This overlap in cognitive processes allows for the development of models that can perform both tasks effectively.

\subsection{Figurative Language Interpretation}

\citet{liu2022testing} are among the first to quantitatively assess the ability of LMs to interpret figurative language. Their \textit{Fig-QA} data set is publicly available\footnote{\url{https://huggingface.co/data sets/nightingal3/fig-qa}} and we discuss the construction of our subcorpus in more detail in Section \ref{sec:data set}. In short, the authors present crowdsourced creative metaphor phrases with two possible interpretations of various LMs and check for which interpretation the model returns the higher probability distribution. The main contribution of \citet{liu2022testing} is the \textit{Fig-QA} task, which consists of 10,256 examples of human-written creative metaphors that are paired in a Winograd schema. The authors also contribute an assessment of various LMs in zero-shot, few-shot and fine-tuned settings on \textit{Fig-QA}. Moreover, their results indicate that overall, LMs fall short of human performance. On a phrase and word level, the authors find that longer phrases are harder to interpret and that metaphors relying on commonsense knowledge concerning objects' volume, height, mass, brightness or colour are easier to interpret. This indicates that bodily modalities seem to facilitate interpretation success. They also show that larger models (i.e. number of parameters) perform better on the task. All of these findings have been reproduced by our experiments.

\citet{chakrabarty2022flute} present FLUTE, a data set of 8,000 figurative NLI instances. Their data set includes the different figurative language categories of metaphor, simile, and sarcasm. In contrast to \textit{Fig-QA}, the authors do not create metaphors in a Winograd scheme as a forced-choice task but create natural language explanations (NLE) using GPT-3 \cite{brown2020language} and human validation. Their experiments with state-of-the-art NLE benchmark models show poor performance in comparison to human performance. The authors do not differentiate the metaphors, similes and sarcastic phrases concerning linguistic features. Moreover, they include a language model in the creation process, which, as far as our study is concerned, introduces a bias to the data set. Hence, we decide to use the \textit{Fig-QA} data set instead of FLUTE.

\subsection{Modelling Embodied Language}

It is generally understood that language is grounded in experience based on interaction with the world \cite{bender2020climbing,bisk2020experience}. Hence, there is an interest to leverage LMs' capabilities in interactions with the environment. For example, \citet{suglia2021embodied} present EmBERT, which attempts language-guided visual task completion. Their model uses a pre-trained BERT stack fused with an embedding for detecting objects from visual input. The model achieves competitive performance on ALFRED, a benchmark task for interpreting instructions \cite{shridhar2020alfred}.

\citet{huang2022language} investigate if LMs know enough embodied knowledge about the world to ground high-level tasks in the procedural planning of instructions for household tasks. For example, the authors pass a prompt, e.g. ``\textit{Step 1: Squeeze out a glob of lotion}'' to a pre-trained LM (e.g. GPT-3) and extract actionable knowledge from its response. Their results indicate that large language models (<10B parameters) can produce plausible action plans for embodied agents.

\paragraph{Embodiment}

In this study, the term \textit{embodiment} relates to cognitive sciences: Humans process a linguistic statement such as ``\textit{to grab an apple}'' using embodied simulations in the brain. Perceptual experiences activate cortical regions that are dedicated to sensory actions and those regions partially reactivate premotor areas to implement, what \citet{barsalou1999perceptual} calls, \textit{perceptual symbols}. Reading of actions words such as \textit{kick} or \textit{lick} is associated with premotor cortex activation responsible for controlling movements for these actions \cite{hauk2004somatotopic}. This effect is diminished by figurative language \cite{schuil2013sentential}. Therefore, a statement such as ``\textit{to grasp the idea}'' does not necessarily rely on premotor cortex simulation. The semantic processing of the linguistic statement is therefore linked to its context and degree of embodiment in the sense that the action can be simulated by a brain in a body \cite{zwaan2014embodiment}. This understanding of the term \textit{embodiment} guides the evaluation of how language models, which do not have a brain in a body, can interpret figurative language phrases with a varying degree of embodied actions.

\section{Statistical Evaluation}
\label{sec:stats}

The review of related works shows that there are abilities of LMs that go beyond mere language generation, e.g. logical reasoning, and action planning. It is unclear how LMs conceptualise actions that humans conceptualise using interaction with the environment. Figurative language acts as a test bed to assess metaphorical conceptualisations since they are grounded in embodied experience and interaction with the environment. We take \citet{liu2022testing}'s findings as a starting point to focus on the effect of embodiment in figurative language interpretation by langauge models of various sizes. 

\subsection{Experimental Framework}

\paragraph{Embodiment Rating and Data Set}
\label{sec:data set}

To assess the effect of embodiment on the task, we discuss the effects of embodiment in semantic processing and introduce the simplification underlying our study through an example. The \textit{Fig-QA} provides the following item:
\begin{quote}

(A) \textit{The pants were as \textcolor{blue}{faded} as ...}

(A.1) \textit{... the memory of pogs}

(A.2) \textit{... the sun in June}

\noindent with the possible interpretations:

(A.I) \textit{they were very faded}

(A.II) \textit{they were bright}
\end{quote}

\noindent The LM is prompted with each combination of sentence completion and interpretation (i.e. A.1+A.I, A.1+A.II, A.2+A.I, A.2+A.II). Notably, \citet{liu2022testing} have shown that the addition of ``\textit{that is to say}'' as a concatenation between metaphorical phrase and interpretation phrase elicits better model performance, hence we also include this prompt in our studies. Subsequently, the prediction scores of the language modelling head (scores for each vocabulary token) are retrieved and the highest probability becomes the LMs choice of interpretation (for more details, see \cite{liu2022testing}). We compare this example with a different \textit{Fig-QA} item:
\begin{quote}
(B) \textit{She \textcolor{orange}{dances} like a ...}

(B.1) \textit{... fairy}

(B.2) \textit{... turtle}

\noindent with the possible interpretations:

(B.I) \textit{she dances well}

(B.II) \textit{she dances poorly}
\end{quote}
\noindent Given our hypothesis that embodiment affects the LMs' ability to interpret these phrases, we score (A) and (B) concerning the embodiment. As a simplification, we limit the rating of embodiment to the actions within the phrase. Every phrase evaluated has at least one word with a score related to an action. Most of the time, these related actions are verbs. Thus, we rate \textcolor{blue}{\textit{faded}} for (A) and \textcolor{orange}{\textit{dances}} for (B) with respect to their relative embodiment. For this scoring, we consult data by \citet{sidhu2014effects}.

In their empirical study, \citet{sidhu2014effects} characterise a dimension of a relative embodiment for verbs. In the construction of the data set, ``\textit{participants were asked to judge the degree to which the meaning of each verb involved the human body, on a 1–7 scale}'' \cite{sidhu2014effects}. Their resulting data set consists of ratings for 687 English verbs. Our hypothesis is that embodiment is a semantic component which affects the interpretation ability of LMs concerning figurative language. With their data set, the authors provide evidence that the meaning of a verb has a semantic component linked to the human body in the lexical processing of that verb. They assume that more robust semantic activation is generated by more embodied verbs \cite{sidhu2014effects}. This provides us with data set we can apply to our experiment on figurative language. Moreover, their experiment provides additional control variables such as the AoA and word length, which have a known effect on lexical processing \cite{colombo2002influence} and are included in our results (Sec. \ref{sec:results}).

At the time of conducting our experiment, \textit{Fig-QA} did only provide the \textit{training} and \textit{development} data, which we will refer to as train \& dev. Hence, we identify all phrases from the train \& dev data set that contain at least one word with an embodiment rating from \citet{sidhu2014effects}. The process of creating the subcorpus with embodiment ratings ($C_{Emb}$) begins by identifying verbs using \textit{spaCy} \cite{honnibal2020spacy}. The lemmatized versions of the verbs for the metaphorical phrases are then matched with embodiment scores, resulting in a subcorpus ($C_{Emb}$) with 1,438 entries. If more than one verb is present in the metaphorical sentence, the average is assigned. We note that, future work will assess whether a different heuristic for treating multiple actions influences our results. Analogously, we construct a subcorpus of the same size with metaphorical phrases that do not contain an embodied verb ($C_{NoE}$). For both subcorpora, we only keep phrases in which the verb is contained in the Winograd pair. The resulting subcorpora statistics are listed in Table \ref{tab:subcorpora} and further examples from the subcorpus are presented in the Appendix in Section \ref{app:examples}. The previous examples (A) and (B) are thus augmented as follows: 
\begin{quote}
(A) \textit{The pants were as \textcolor{blue}{faded} as ...}

\noindent Embodiment Rating: 2.36

(B) \textit{She \textcolor{orange}{dances} like a ...}

\noindent Embodiment Score: 6.50 
\end{quote}

With the annotated \textit{Fig-QA} subcorpus $C_{Emb}$ we now turn to the models we select to assess whether there is a correlation between embodiment score and LM task performance.


\begin{table*}[]
\centering
\resizebox{\textwidth}{!}{%
\begin{tabular}{lllr}
\textbf{Label} & \textbf{Source}           & \textbf{Description}                                                  & \textbf{Number of entries} \\ \hline
$C_{Liu}$ & Fig-QA test      & All phrases from the Fig-QA test set                         & 1,146             \\
$C_{Emb}$  & Fig-QA train/dev & Phrases that have at least one action with embodiment rating & 1,438             \\
$C_{NoE}$  & Fig-QA train/dev & Phrases that do not have an action with embodiment rating    & 1,438            
\end{tabular}
}
\caption{\label{tab:subcorpora} $C_{Liu}$ is 100\% of the \textit{Fig-QA} test set and 11\% of the entire \textit{Fig-QA} data set \cite{liu2022testing}. Our selected subsets $C_{Emb}$ and $C_{NoE}$ are mutually exclusive and each composes 14\% of the entire \textit{Fig-QA} data set.}
\end{table*}

\paragraph{Hypotheses}
The main hypothesis for the statistical evaluation can be summarized as follows:
\begin{enumerate}
    \item[1.] There is a correlation between the LMs' interpretation capabilities of metaphors and the amount of embodiment of the verbs within those metaphorical phrases.
\end{enumerate}

Intuitively, more embodied actions such as \textit{kick}, \textit{move} or \textit{eat} are much more concrete, shorter and basic, when compared to \textit{resonate}, \textit{compartmentalise} or \textit{misrepresent}. Therefore, the analysis of embodied actions must take into account factors such as concreteness, AoA, word length and word frequency. Moreover, common metaphors are conventional and more lexicalised. Consequently, they might simply be more embodied and the effect of embodied verbs might stem from the fact that these verbs are more concrete in the context that they are presented. Hence, the first hypothesis should not stand alone, but will be evaluated along with two additional null hypotheses:

\begin{enumerate}[resume]
  \item[1.I] There is no correlation between the LMs' interpretation capabilities of metaphors and the amount of concreteness of the verbs within those metaphorical phrases irrespective of their embodiment rating.
\end{enumerate}

In our evaluation, the concreteness of a word in its context will be scored using an open-source predictor\footnote{\url{https://github.com/armandrotaru/TeamAndi-CONcreTEXT}} based on distributional models and behavioural norms explained in \cite{rotaru2020andi}. Details of the concreteness scoring with the predictor have been summarized in Sec. \ref{sec:concret}. Concreteness ratings are often subjective ratings \cite{brysbaert2014concreteness} or determined by other low-level features, such as AoA, word frequency and word length \cite{rotaru2020andi}. To isolate the effect of embodiment, we add the second null hypothesis:

\begin{enumerate}[resume]
  \item[1.II] There is no correlation between the LMs' interpretation capabilities of metaphors and other linguistic features, such as AoA, word frequency and word length.
\end{enumerate}

For AoA we obtain scores for each of the actions from \cite{kuperman2012age} and for word frequency from \cite{van2014subtlex}. Together with word length and embodiment score we test for variance inflation to respond to 1.II.

\paragraph{Model Selection}

The selection of our models is based on three criteria: First, we want to reproduce the results by \cite{liu2022testing} having a comparable measure. Second, we want to check whether the effect generalises to other large LMs. Third, we want a variation of different model sizes to account for varying performance on the task as a result of model size. For the latter two criteria, we start with the smallest available models of each type and check intermediate model sizes. We do not consider it necessary to check whether or not scaled, largest versions of each model perform better on the task since this is a general property of LMs \cite{brown2020language,srivastava2022beyond}.

In the original \textit{Fig-QA} study, the authors examine three transformer-based LMs with different parameter sizes: GPT-2 \cite{radford2019language}, GPT-3 \cite{brown2020language} and GPT-Neo \cite{gpt-neo}. To reproduce the results by \citet{liu2022testing}, we include GPT-2, GPT-3, GPT-Neo LMs and add OPT LMs \cite{zhang2022opt}. An overview of the models and their specifications is shown in Table \ref{tab:models}. Notably, we want to correlate whether the type and number of parameters play a role when it comes to performance concerning the embodiment. Hence, we include pairs of models from each type that are small (<1 billion) and medium to large (>1 billion) in their number of parameters. 

\begin{table}[]
\tiny
\centering
\resizebox{.48\textwidth}{!}{%
\begin{tabular}{ccc}
\textbf{Label} & \textbf{\begin{tabular}[c]{@{}c@{}}\#Parameters\\  in Millions\end{tabular}} & \textbf{Provider} \\ \hline
\begin{tabular}[c]{@{}c@{}}GPT-3\\ (small)\end{tabular} & $\sim$350 & OpenAI \\
\begin{tabular}[c]{@{}c@{}}GPT-3\\ (large)\end{tabular} & $\sim$175,000 & OpenAI \\ \hline
\begin{tabular}[c]{@{}c@{}}OPT\\ (small)\end{tabular} & 350 & Facebook \\
\begin{tabular}[c]{@{}c@{}}OPT\\ (medium)\end{tabular} & 13,000 & Facebook \\ \hline
\begin{tabular}[c]{@{}c@{}}GPT-Neo\\ (small)\end{tabular} & 125 & EleutherAI \\
\begin{tabular}[c]{@{}c@{}}GPT-NeoX\\ (medium)\end{tabular} & 20,000 & EleutherAI \\ \hline
\begin{tabular}[c]{@{}c@{}}GPT-2\\ (small)\end{tabular} & 355 & OpenAI \\
\begin{tabular}[c]{@{}c@{}}GPT-2 XL\\ (medium)\end{tabular} & 1,500 & OpenAI \\ \hline
\end{tabular}%
}
\caption{\label{tab:models} Different model types and parameter numbers have been selected for the evaluation. For each type, we have selected a pair of smaller and medium to large model version.}
\end{table}

\subsection{Methodology}

We apply the same methodology of evaluation as \citet{liu2022testing}. In our zero-shot setting, each pretrained LM is prompted with the metaphor sentences combined with one of the interpretation sentences, concatenated with \textit{that is to say}. For \textit{OpenAI} models, the API provides the log probabilities per token as \texttt{logprob} return value. We access all other models using \url{huggingface.co} and its \texttt{transformer} library. To create the same evaluation metric as for results by \cite{liu2022testing}, we follow \cite{tunstall2022natural} and implement a function that returns the \texttt{logprob} based on the prediction scores of the language modelling head. All code and data is publicly available\footnote{\url{https://osf.io/puhxb/?view_only=15933a2da0a14f07834ba1d479ce9c43}}.

\paragraph{Reproduction and Suffix Prompting}
In an initial experiment, we reproduce the same experiment by \citet{liu2022testing}, but instead of performing the zero-shot classification on the test set, we evaluate the performance on $C_{Emb}$ and $C_{NoE}$ with and without suffix prompting (\textit{that is to say}). This allows us to compare our data set against the baseline. The result indicates that GPT-3 models perform slightly worse on our subcorpora, but all conditions benefit from the suffix prompting (see Appendix \ref{sec:appendixA}). Since both $C_{Emb}$ and $C_{NoE}$ are more difficult for GPT-3, we can rule out that this effect stems solely from the embodiment component present in $C_{Emb}$, which is not present in $C_{NoE}$. Moreover, we adopt the suffix prompting for all further experiments.

\paragraph{Concreteness Scoring}
\label{sec:concret}

To determine the concreteness of a verb in context, \cite{rotaru2020andi} built a predictor based on a combination of distributional models, together with behavioural norms. We adopt the same settings and model choice as presented by the author, but exclude the word frequency behavioural norm, as we investigate it as a separate feature. We evaluate our predictor on the same English test set of the \textit{CONcreTEXT} task at \textit{EVALITA2020} \cite{gregori2020concretext} and receive a mean Spearman correlation of 0.87, which is in line with \cite{rotaru2020andi}. The context-dependent models used in the predictor include ALBERT \cite{lan2019albert}, BERT  and GPT-2.

\paragraph{Statistical Tests}
For each model, we obtain its performance on the data set with a binary scoring of each figurative phrase as being correctly or incorrectly identified. We correlate this series of binary values with the continuous variable of embodiment ratings by calculating the point biserial correlation coefficient and the associated p-value. Moreover, we assess various other language features to isolate any effect of embodiment. As described in previous sections and based on the work by \citet{liu2022testing,sidhu2014effects,colombo2002influence}, we test for the effects of word concreteness, AoA, word frequency and word length. This analysis includes an assessment of the amount of multicollinearity within the regression variables by determination of the variance inflation factor (VIF). Moreover, we conduct linear regressions for all models (and all sizes) with respect to their task performance and the features: embodiment score, AoA, word frequency and word length. For these linear regressions, we include those with and without the embodiment score feature in order to assess whether this feature contributes to a higher coefficient of determination ($R^2$).

\section{Results}
\label{sec:results}

\begin{figure}
    \centering
    \includegraphics[width=.48\textwidth]{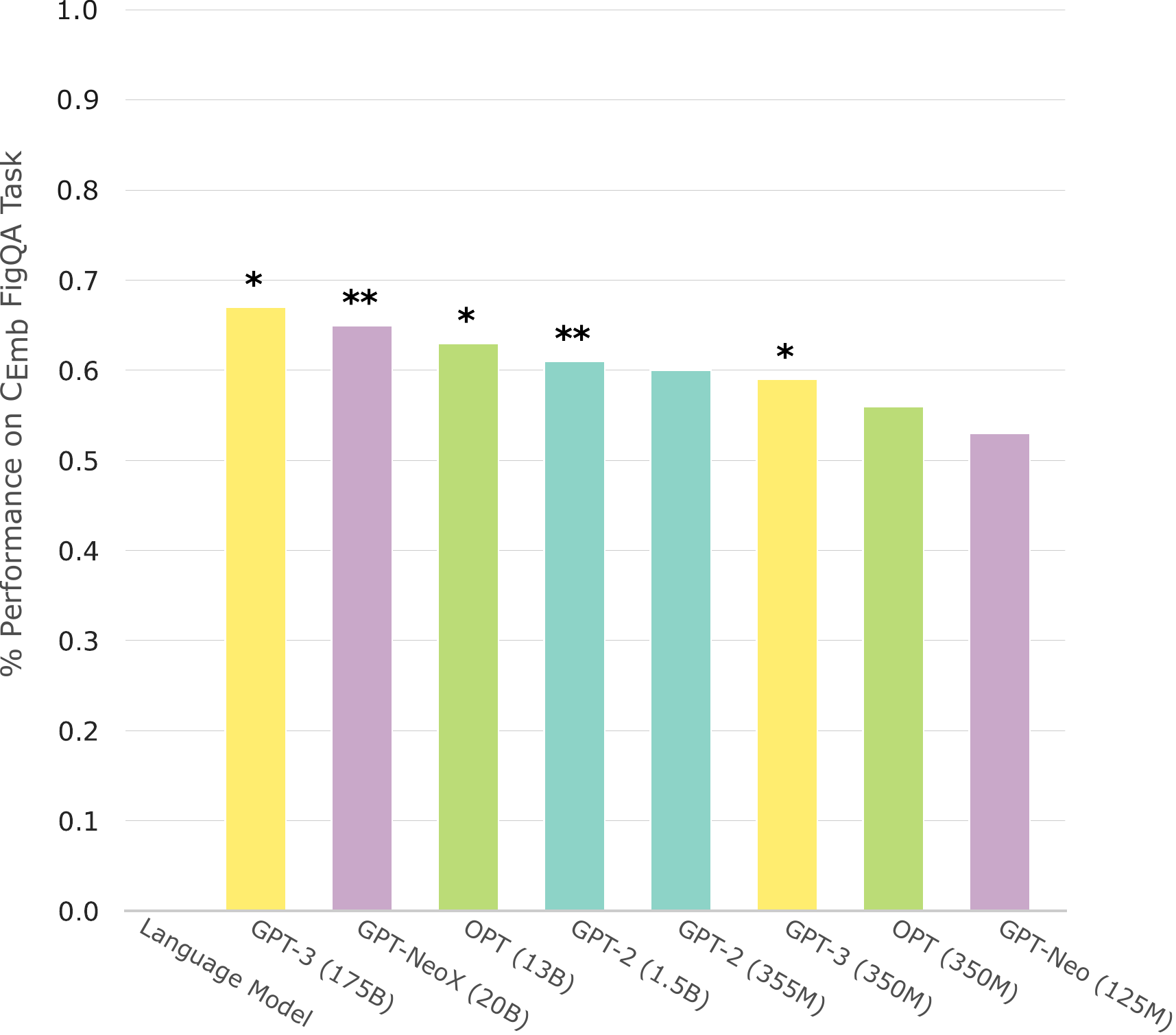}
    \caption{Performance of four language models in two size-variations on $C_{Emb}$. Significant results of the point biserial correlation between embodiment score and model performance are marked for $p<0.05$ with {*} and for $p<0.01$ with {*}{*}. Colours correspond to the same model type, and x-labels provide model size.}
    \label{fig:result_models}
\end{figure}

\paragraph{Embodiment Correlation}
The results of all models are listed in Table \ref{tab:results} and visualized in Figure \ref{fig:result_models}. Overall, for each pair of small and larger models, the larger models always perform better on the interpretation task than the smaller version of the model. Moreover, \textit{all} larger versions of the models show a significant correlation ($p<0.05$) between the embodiment rating and task performance. In two instances, GPT-NeoX (20B) and GPT-2 (1.5B), the  $p$ value is $<0.01$. In the case of GPT-3, both model variants show a significant correlation. In all correlations, the coefficient is positive, albeit small (<0.1), which indicates that embodiment has a positive effect on task performance. All smaller models (except for GPT-3 with 350M parameters) do not show a significant correlation between embodiment score and task performance. 

\begin{table}[]
\resizebox{.48\textwidth}{!}{%
\begin{tabular}{ccccc}
\hline
\textbf{Model} & \textbf{\begin{tabular}[c]{@{}c@{}}Accuracy on\\ $C_{Emb}$\end{tabular}} & \textbf{p-Value} & \textbf{\begin{tabular}[c]{@{}c@{}}Correlation\\ coefficient\end{tabular}} & \textbf{\begin{tabular}[c]{@{}c@{}}{*} p \textless  .05\\ {*}{*} p \textless  .01\end{tabular}} \\ \hline
\begin{tabular}[c]{@{}c@{}}GPT-3\\ (small)\end{tabular} & 0.594 & 0.018 & 0.062 & \textbf{*} \\
\begin{tabular}[c]{@{}c@{}}GPT-3\\ (large)\end{tabular} & 0.667 & 0.034 & 0.056 & \textbf{*} \\ \hline
\begin{tabular}[c]{@{}c@{}}OPT\\ (small)\end{tabular} & 0.561 & 0.206 & 0.033 &  \\
\begin{tabular}[c]{@{}c@{}}OPT\\ (medium)\end{tabular} & 0.627 & 0.034 & 0.056 & \textbf{*} \\ \hline
\begin{tabular}[c]{@{}c@{}}GPT-Neo\\ (small)\end{tabular} & 0.535 & 0.399 & 0.022 &  \\
\begin{tabular}[c]{@{}c@{}}GPT-NeoX\\ (medium)\end{tabular} & 0.648 & 0.005 & 0.073 & \textbf{**} \\ \hline
\begin{tabular}[c]{@{}c@{}}GPT-2\\ (small)\end{tabular} & 0.597 & 0.158 & 0.037 &  \\
\begin{tabular}[c]{@{}c@{}}GPT-2 XL\\ (medium)\end{tabular} & 0.606 & 0.009 & 0.069 & \textbf{**} \\ \hline
\end{tabular}%
}
\caption{\label{tab:results} Experimental results of all model pairs (small and larger versions) on the $C_{Emb}$ corpus. The last column marks significant results of the point biserial correlation between embodiment score and model performance for $p<0.05$ with {*} and for $p<0.01$ with {*}{*}.}
\end{table}

\paragraph{Concreteness}
Using the concreteness-in-context predictor, we provide a concreteness value for each verb in $C_{Emb}$ and correlate those predictions with all models' performance. As a result, there is no significant correlation between the concreteness of the action word in its context and the performance of the LM on the interpretation (results in Appendix \ref{app:conc}). We do not reject hypothesis 1.I.

\paragraph{Regression Analysis}
The linear regressions for all models and model sizes, both with and without the feature of embodiment score, revealed that the coefficient of determination ($R^2$) was consistently higher for regressions that included the embodiment score feature. Furthermore, for cases without the embodiment feature, none of the other variables, such as Age of Acquisition (AoA), word frequency, or word length, showed a significant correlation with task performance. Related results and figures 
 are available in Section \ref{app:linreg} of the Appendix.

\paragraph{Variance Inflation}

\begin{figure}
    \centering
    \includegraphics[width=.48\textwidth]{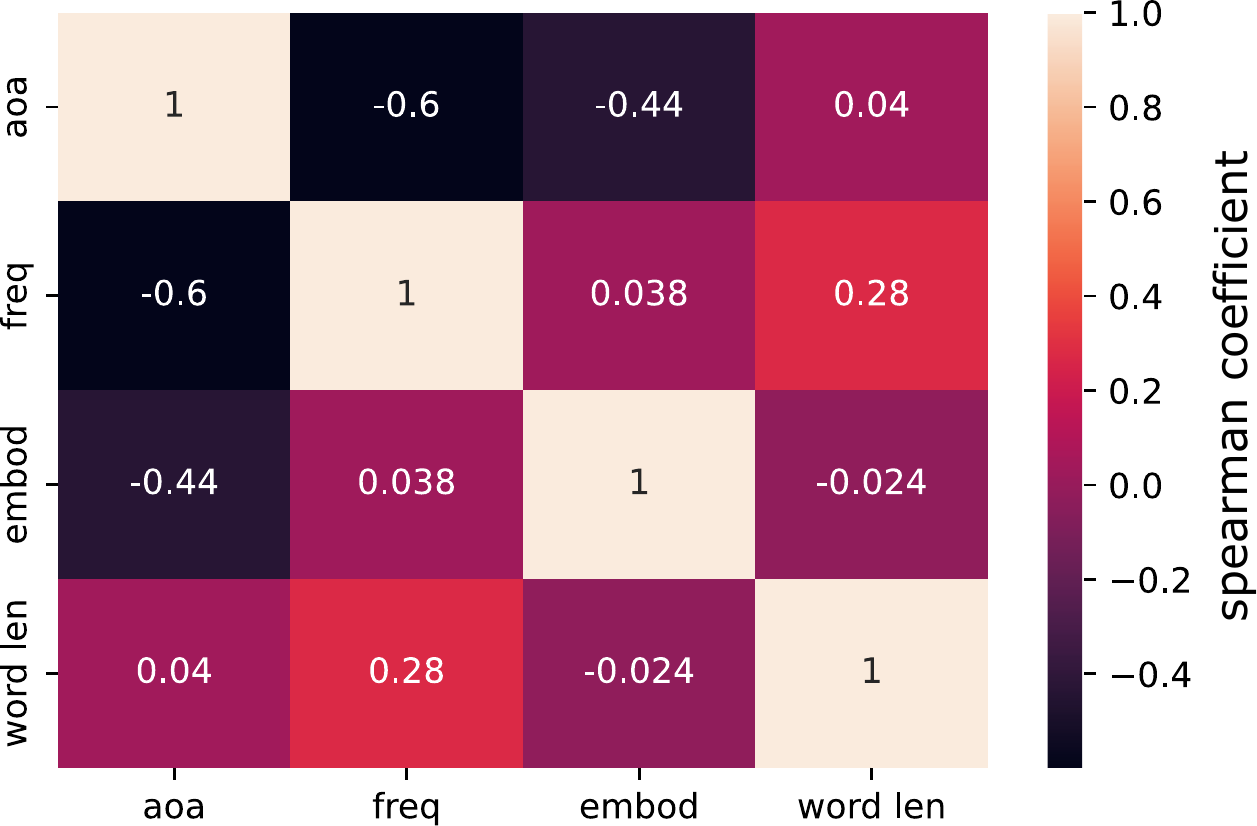}
    \caption{\label{fig:corr}Pairwise correlation for the variables of AoA, word frequency, embodiment score and word length. Positive numbers indicate that increase in one variable correlates with an increase in the other and analogously for negative numbers with a decrease. Value 1 is the perfect correlation of the variable with itself. \texttt{aoa}: Age of acquisition, \texttt{freq}: Word frequency, \texttt{embod}: Embodiment score and \texttt{word len}: word length.}
    \label{fig:results}
\end{figure}

Pairwise correlations between AoA, word frequency, embodiment score and word length are visualized in Figure \ref{fig:corr}. Intuitively, frequency and AoA are expected to be correlated with each other, because words that are acquired much later in life are often less frequently used words, as they tend to be more complex or specific words. The multicollinearity test through VIF is presented in Table \ref{tab:vif}. All factors are close to 1.0, which indicates that there is no multicollinearity among predictor variables (if the VIF is between 5 and 10, multicollinearity is likely to present) \cite{james2013introduction}. Given that there is no multicollinearity between embodiment score and other linguistic features, we do not reject hypothesis 1.II.

\begin{table}[]
\resizebox{.48\textwidth}{!}{%
\begin{tabular}{ccccc}
AoA & \begin{tabular}[c]{@{}c@{}}Word\\ Frequency\end{tabular} & \begin{tabular}[c]{@{}c@{}}Embodiment\\ Rating\end{tabular} & \begin{tabular}[c]{@{}c@{}}Word \\ Length\end{tabular} & Constant \\ \hline
1.610 & 1.345 & 1.326 & 1.017 & 86.387
\end{tabular}%
}
\caption{\label{tab:vif}VIF from the four features. Constant denotes the intercept provided for the VIF. A factor close to 1 indicates no correlation with values above 4 regarded as moderate correlation.}
\end{table}

\subsection{Interpretation}

The results of the correlation analysis indicate that embodiment affects the LMs' ability to interpret figurative language when the LM achieves a certain level of performance, which depends on the size of the model. The correlation coefficient is positive in all significant cases and those significant correlations occur in all larger (>1B parameter) model versions. Since task performance increases with model size, the effect of embodiment becomes more apparent through more successful interpretations in better-performing models. The fact that concreteness, AoA, word length and word frequency do not inflate this effect, shows that the embodiment rating is not an arbitrary construct that implicitly models another linguistic feature.

There are slight differences in the model types when it comes to the performance of the models. For example, GPT-3 shows a significant correlation for the effect of an embodiment for both, the small (350M) and large (175B) model sizes. Yet, this effect does not occur in the small GPT-2 (355M) model, but in the large GPT-2 (1.5B) version. Notably, \textit{OpenAI} does not explicitly list the \textit{Ada} model with 350M, but its performance ranks close with 350M versions on various tasks \cite{brown2020language}. Hence, this difference has only limited relevance. Nonetheless, we assume that the effect is correlating with model size and that a reliable effect can be seen in larger models with a parameter number of over 1 billion.

\section{Conclusion}
\label{sec:concl}

\subsection{Contribution to the Field}

We successfully reproduce results that are in line with \cite{liu2022testing}. Moreover, we provide a subcorpus with ratings of an embodiment for the \textit{Fig-QA task}. We identify the contribution of embodied verbs to LMs' ability to interpret figurative language.  To the best of our knowledge, this study is the first to provide evidence that the psycho-linguistic norm of the perceived embodiment has been investigated in an NLP task for LMs.

\subsection{Discussion}

Benchmarks, such as BIG-bench \cite{srivastava2022beyond}, show that different types and sizes of LMs can be evaluated on many different tasks to identify potential shortcomings or limitations. This paper takes an entirely different approach by zeroing in on a particular task, which has been augmented with a specific semantic evaluation (embodiment ratings of actions), to highlight how difficult tasks, such as figurative language interpretation, benefit not only from model size but from specific embodied semantics. 

Figurative language is difficult for LMs because its interpretation is often not conveyed directly by the conventional meaning of its words. Human NLU is embodied and grounded by physical interaction with the environment \cite{di2018linguistic}. Consequently, it could be expected that LMs struggle when the interpretation of figurative language depends on a more embodied action. Yet, the opposite has been shown as more embodied concepts are more lexicalised and larger LMs can interpret them better in figurative language. Hence, our study provides valuable insight that raises the question of whether this effect is limited to figurative language or translates to other NLU tasks for LMs. 

\subsection{Limitations and Future Works}

The current results are limited to one specific figurative language task (\textit{Fig-QA)}. In future work, we aim to test whether our hypothesis holds for other figurative language interpretation tasks, such as those by \cite{chakrabarty2022flute,stowe-etal-2022-impli}. Moreover, we want to assess BIG-bench \cite{srivastava2022beyond} performances on various other tasks concerning embodiment scoring and see whether the bias can be detected in tasks other than figurative language interpretation.  

The statistical evaluation has attempted to measure many different linguistic dimensions, e.g. AoA, word length, word frequency and concreteness in context. Empirically, this indicates that the effect of embodiment is not simply explainable by other factors. Theoretically, we argue that this correlation can be causally explained through the lexicalisation of conventional metaphors. We simplify conventionality by assuming that word frequency and age of acquisition (AoA) are indicators of conventionality, i.e. more frequent words and words that are acquired early in life are more conventional. Nonetheless, a thorough explanation of the effect of embodiment on LMs' capabilities for language tasks requires many more studies.

\paragraph{Ethical Consideration} It should be noted that a key component of the experiment is built from \cite{sidhu2014effects} with their ratings of relative embodiment. For their study, the authors have sampled data exclusively from ($N$=67, 57 female) ``\textit{graduate students at the University of Calgary who participated in exchange for bonus credit in a psychology course, had a normal or corrected-to-normal vision, and reported English proficiency}'' \cite{sidhu2014effects}. Even though \textit{embodiment} is supposed to be a general, human experience, the pool of participants is relatively homogeneous (mostly female, educated and presumably able-bodied). A broader and more diverse set of ratings, specifically concerning differently-abled participants and cultural backgrounds should be targeted. 

\paragraph{Computing Cost} All model inferences (except \textit{OpenAI}) have been conducted on University servers with 8x \textit{NVIDIA RTX} A6000 (300 W). Each experiment for each model lasted at most 10min with full power consumption. A conservative estimate of 2,400 W (8 GPUs x 300 W) for 20 experiments results in a power consumption of at most 8 kWh, which equals emission of at most $\sim$3.5 kg $CO_2$ for all experiments with model inference.

\paragraph{Data and Artefact Usage}
Existing artefacts used in this research are attributed to their creators and their consent has been acquired before the studies. This concerns the \textit{embodiment ratings} by \citet{sidhu2014effects}, the \textit{Fig-QA} corpus by \citet{liu2022testing} and the \textit{concreteness predictor} by \citet{rotaru2020andi}.

\section*{Acknowledgements}

We would like to thank the Pexman Language Processing Lab (University of Calgary) for sharing the embodiment ratings. We would also like to thank Team Andi for publishing and providing their CONcreTEXT submission. Without model access and hosting by \textit{Huggingface} and \textit{OpenAI} the study would not have been possible. We would also like to thank Marianna Bolognesi and Stefan Riegl for their valuable feedback. Moreover, we thank the three ARR reviewers for their valuable feedback.

\bibliography{anthology,custom}
\bibliographystyle{acl_natbib}

\appendix

\section{Appendix: Further samples form $C_{Emb}$} \label{app:examples}

Further samples from $C_{Emb}$ are provided in Table \ref{tab:app_a}. The table depicts a selection of linguistic examples of more embodied and less embodied metaphors. Originally, these samples are form the \textit{Fig-QA} data set by \citet{liu2022testing}, providing the entries for the columns \textit{Figurative Phrase} and \textit{Target Interpretation}. The \textit{Embod. Score} is the embodiment rating of the action identified in the respective column and taken from \citet{sidhu2014effects}. A lower embodiment score indicates that the action has been rated as ``less embodied''. In Table \ref{tab:app_a}, none of the language models shows a statistically significant correlation between the variables ($\alpha<0.05$).

\begin{table*}[t]
\small
\centering
\begin{tabular}{llrl}
\textbf{Figurative Phrase}                                & \textbf{Target Interpretation}                   & \multicolumn{1}{l}{\textbf{Embod. Score}} & \textbf{Action}     \\ \hline 
                                                 &                                         & \multicolumn{1}{l}{}             &            \\[-1.5ex]
The chihuahua believes it is a wolf              & The small dog thinks it is undefeatable & 2.83                             & believe    \\
The chihuahua believes it is a lap blanket       & The small dog always stays on your lap  & 2.83                             & believe    \\[-1.5ex]
                                                 &                                         & \multicolumn{1}{l}{}             &            \\
He knew her like a sister                        & He knew her very well.                  & 3.00                             & know      \\
He knew her like a stranger                      & He didn't know her well.                & 3.00                             & know      \\[-1.5ex]
                                                 &                                         & \multicolumn{1}{l}{}             &            \\
He guided her like a lighthouse.                 & He was a good guide.                    & 3.61                             & guide      \\
He guided her like a broken GPS.                 & He was a terrible guide.                & 3.61                             & guide      \\[-1.5ex]
                                                 &                                         & \multicolumn{1}{l}{}             &            \\
The argument appears as a crystal clear spring   & The argument makes sense                & 3.90                              & appear     \\
The argument appears as a muddy rut              & The argument is senseless               & 3.90                              & appear     \\[-1.5ex]
                                                 &                                         & \multicolumn{1}{l}{}             &            \\
the movie raised your spirits to heaven          & The movie was uplifting.                & 4.29                             & raise      \\
the movie raised your spirits to the ocean floor & The movie was depressing.               & 4.29                             & raise      \\[-1.5ex]
                                                 &                                         & \multicolumn{1}{l}{}             &            \\
It was buried as deep as an oil well             & It was buried deep                      & 4.87                             & bury       \\
It was buried as deep as a bathtub               & It was not buried deep                  & 4.87                             & bury       \\[-1.5ex]
                                                 &                                         & \multicolumn{1}{l}{}             &            \\
The reporter wrote like a monkey on crack        & The reporter wrote badly                & 5.19                             & write      \\
The reporter wrote like Hemingway                & The reporter wrote well                 & 5.19                             & write      \\[-1.5ex]
                                                 &                                         & \multicolumn{1}{l}{}             &            \\
He should be cooking for Gordon Ramsay           & His cooking is good  	               & 5.47                             & cook      \\
He should be cooking for McDonalds               & His cooking is bad                      & 5.47                             & cook      \\[-1.5ex]
                                                 &                                         & \multicolumn{1}{l}{}             &            \\
She sings like a nightingale                     & She sings beautifully                   & 6.03                             & sing       \\
She sings like an angry crow                     & She sings horribly                      & 6.03                             & sing       \\[-1.5ex]
                                                 &                                         & \multicolumn{1}{l}{}             &            \\
The food tasted like eating a mother's love      & The food tasted amazing                 & 6.26                            & eat, taste \\
The food tasted like eating the bottom of a shoe & The food tasted disgusting              & 6.26                            & eat, taste \\[-1.5ex]
                                                 &                                         & \multicolumn{1}{l}{}             &            \\
He could sprint like the wind                    & He was fast.                            & 6.46                             & sprint     \\
He could sprint like a tortoise                  & He was slow.                            & 6.46                             & sprint    
\end{tabular}
\caption{\label{tab:app_a} Linguistic examples of more embodied and less embodied metaphors, sampled from $C_{Emb}$ (derived from \citet{liu2022testing}). Every pair of sentences is presented with both target sentences to each model. Embodiment scores retrieved from \citet{sidhu2014effects}.}
\end{table*}

%

\section{Appendix: Concreteness Correlation}
\label{app:conc}
Table \ref{app:concorr} shows the results of the the point biserial correlation between the concreteness of action in context and performance of LM on the interpretation of figurative language phrases ($C_{Emb}$). \texttt{pointbiserial} of the \textit{scipy stats} package for \textit{Python} has been used to determine the correlation. This function uses a t-test with n-1 degrees of freedom. The value of the point-biserial correlation has been calculated from: 
\begin{equation}
   r_{\text{pb}} = \frac{{(\bar{Y_1} - \bar{Y_0})}}{{s_y}} \sqrt{\frac{{N_1 N_2}}{{N(N-1)}}}
\end{equation}

With $Y_0$ and $Y_1$ as the means of the metric observations; $N_1$ and $N_2$ as the number of observations; $N$ as the total number of observations and $s_y$ as the standard deviation of all the metric observations.

\begin{table*}[]
\centering
\small
\begin{tabular}{l|cccccccc}
\textbf{Model} & \textbf{GPT-3} & \textbf{GPT-3} & \textbf{OPT} & \textbf{OPT} & \textbf{GPT-Neo} & \textbf{GPT-NeoX} & \textbf{GPT-2} & \textbf{GPT-2 XL} \\
Parameters     & 350M           & 175B           & 350M         & 13B          & 125M             & 20B               & 355M           & 1.5B              \\
Correlation    & -0.016         & -0.039         & -0.037       & -0.032       & -0.020           & -0.026            & -0.036         & -0.003            \\
p-value        & 0.554          & 0.139          & 0.163        & 0.227        & 0.448            & 0.318             & 0.176          & 0.921            
\end{tabular}
\caption{\label{app:concorr}Results of the point biserial correlation between the concreteness of action in context and performance of LM on the interpretation of figurative language phrases ($C_{Emb}$). None of the LMs shows a statistically significant correlation between the variables ($\alpha<0.05$).}
\end{table*}

\section{Appendix: Reproduction and Suffix Prompting}
\label{sec:appendixA}
Table \ref{tab:appCorr} presents the results of the zero-shot performance of the GPT-3 models \textit{Ada} ($\sim$350M parameters) and \textit{Davinci} ($\sim$175B parameters) on the different corpora (Table \ref{tab:subcorpora}) with respect to the suffix \textit{that is to say}. On all data sets ($C_{Liu}$, $C_{Emb}$, $C_{NoE}$ and $C_{Emb}+C_{NoE}$) performance of both models is better if the suffix is provided.

\begin{table*}[]
\centering
\begin{tabular}{lllllll}
 &  & \multicolumn{2}{l}{GPT-3 (small)} &  & \multicolumn{2}{l}{GPT-3 (large)} \\ \cline{3-4} \cline{6-7} 
Corpus &  & w/ suffix & w/o suffix &  & w/ suffix & w/o suffix \\ \cline{1-1} \cline{3-4} \cline{6-7} 
$C_{Liu}$ & & 0.601 & 0.591 &  & - & 0.684 \\
$C_{Emb}$ & & 0.594 & 0.577 &  & 0.667 & 0.661 \\
$C_{NoE}$ & & 0.583 & 0.572 &  & 0.661 & 0.659 \\
$C_{Emb}+C_{NoE}$ &  & 0.591 & 0.574 &  & 0.665 & 0.660
\end{tabular}%
\caption{\label{tab:appCorr}Comparing the zero-shot performance of the GPT-3 models \textit{Ada} ($\sim$350M parameters) and \textit{Davinci} ($\sim$175B parameters) on the different corpora (Tab. \ref{tab:models}). The comparison includes  the variable \textit{that is to say} suffix prompting.}
\end{table*}

\section{Appendix: Linear Regression Data}
\label{app:linreg}

\begin{figure*}[t!]
    \centering
    \includegraphics[width=.8\textwidth]{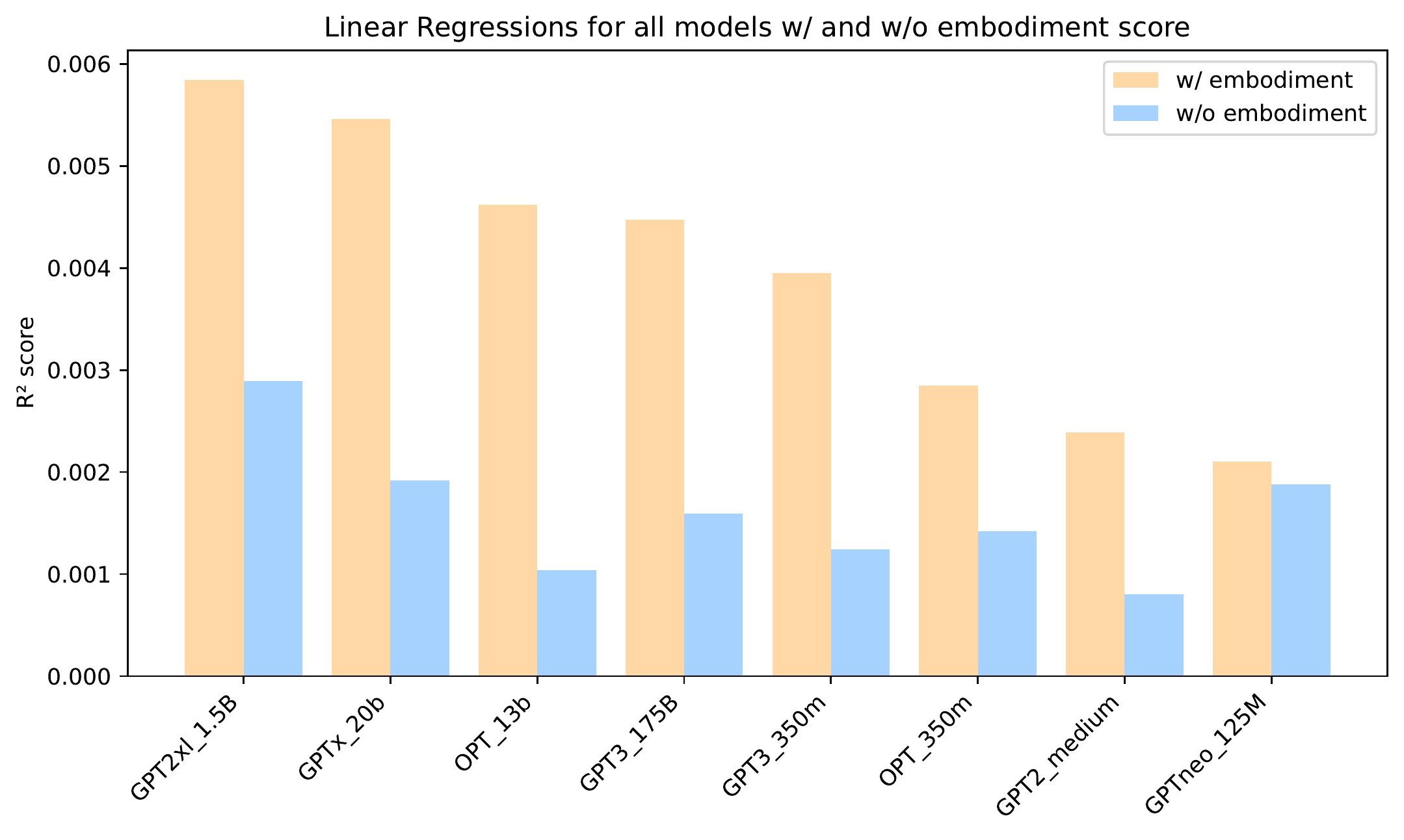}
    \caption{Coefficients of determination ($R^2$) for all models and all model sizes through linear regression with (orange) and without (blue) embodiment score as feature. For all regressions, the features Age of Acquisition, word frequency and word length have been included. For all models and sizes, the $R^2$ value is lower when embodiment is excluded as a feature. This is in line with the VIF analysis. Details of these linear regressions are exemplified in the results for \textit{GPT3\_350m} in Table \ref{tab:wEmbod} (with embodiment score feature) and in Table \ref{tab:woEmbod} (without embodiment score feature).
    \label{app:r2s}}
\end{figure*}

 In addition to the VIF, we performed two linear regression analyses for each of the models (and sizes). These results are visualized in Figure \ref{app:r2s}. The first includes all features (embodiment score, Age of Acquisition, word frequency and word length). The second excludes the embodiment score and shows a lower coefficient of determination ($R^2$) for all regressions.  Details of these linear regressions are exemplified in the results for \textit{GPT3\_350m} in Table \ref{tab:wEmbod} (with embodiment score feature) and in Table \ref{tab:woEmbod} (without embodiment score feature).

\begin{table*}[t]
\centering
\small
\begin{tabular}{lllllllll}
\textbf{feature}  & \textbf{coef} & \textbf{se} & \textbf{T} & \textbf{p-val}   & $\mathbf{R^2}$ & \textbf{adj\_r2} & \textbf{CI{[}2.5\%{]}} & \textbf{CI{[}97.5\%{]}} \\ \hline
Intercept         & 0.45075       & 0.12647     & 3.56405    & 0.00038          & 0.00393              & 0.00115          & 0.20266                & 0.69884                 \\
embodiment rating & 0.02906       & 0.01449     & 2.00613    & \textbf{0.04503} &                      &                  & 0.00064                & 0.05748                 \\
freq-rating       & -0.00000      & 0.00000     & -0.07397   & 0.94104          &                      &                  & -0.00000               & 0.00000                 \\
aoa-rating        & 0.00163       & 0.01278     & 0.12763    & 0.89846          &                      &                  & -0.02344               & 0.02671                 \\
len-rating        & 0.00073       & 0.01385     & 0.05249    & 0.95814          &                      &                  & -0.02645               & 0.02790                
\end{tabular}
\caption{\label{tab:wEmbod}Linear Regression results with all features including embodiment score for GPT-3 (\textit{ada}). \textit{coef}: regression coefficients, \textit{se}: standard errors, \textit{T}: T-values, \textit{p-val}: p-values, \textit{r2}: coefficient of determination ($R^2$), \textit{$adj_{r2}$}: adjusted $R^2$, \textit{CI{[}2.5\%{]}}: lower confidence intervals, \textit{CI{[}97.5\%{]}}: upper confidence intervals.}
\end{table*}

\begin{table*}[t]
\centering
\small
\begin{tabular}{lllllllll}
\textbf{Feature} & \textbf{coef} & \textbf{se} & \textbf{T} & \textbf{pval} & $\mathbf{R^2}$ & \textbf{adj\_r2} & \textbf{CI{[}2.5\%{]}} & \textbf{CI{[}97.5\%{]}} \\ \hline
Intercept        & 0.6659        & 0.0671      & 9.9174     & 0.00000       & 0.00114              & -0.00095         & 0.5342                 & 0.7976                  \\
freq-rating      & -0.00000      & 0.00000     & -1.0400    & 0.29852       &                      &                  & -0.00000               & 0.00000                 \\
aoa-rating       & -0.00994      & 0.01142     & -0.8700    & 0.38434       &                      &                  & -0.03234               & 0.01246                 \\
len-rating       & -0.00227      & 0.01379     & -0.1648    & 0.86916       &                      &                  & -0.02931               & 0.02477                 \\
len-rating       & 0.00073       & 0.01385     & 0.05249    & 0.95814       &                      &                  & -0.02645               & 0.02790                
\end{tabular}
\caption{\label{tab:woEmbod}Linear Regression results with all features excluding embodiment score for GPT-3 (\textit{ada}). \textit{coef}: regression coefficients, \textit{se}: standard errors, \textit{T}: T-values, \textit{p-val}: p-values, \textit{r2}: coefficient of determination ($R^2$), \textit{$adj_{r2}$}: adjusted $R^2$, \textit{CI{[}2.5\%{]}}: lower confidence intervals, \textit{CI{[}97.5\%{]}}: upper confidence intervals}.
\end{table*}

\end{document}